\documentclass[1p]{elsarticle}
\usepackage{hyperref}
\usepackage{amssymb}
\usepackage{multirow}
\usepackage{amsmath}
\usepackage{synttree}
\usepackage{caption}
\usepackage{xcolor}
\usepackage[T1]{fontenc}

\journal{Expert Systems with Applications}

\biboptions{authoryear}

\begin{document}

\begin{frontmatter}

\title{Semi-automatic Generation of Multilingual Datasets for Stance
Detection in Twitter\footnote{Please cite this paper as: Elena Zotova, Rodrigo
Agerri, German Rigau. Semi-automatic Generation of Multilingual Datasets for
Stance Detection in Twitter, Expert Systems with Applications (2021), 170.
\url{https://doi.org/10.1016/j.eswa.2020.114547}. \copyright 2021.}
\footnote{This manuscript is made available under the CC-BY-NC-ND 4.0 license
\url{http://creativecommons.org/licenses/by-nc-nd/4.0/}. Paper submitted 29
July 2020, Revised 4 December 2020, Accepted 24 December 2020.}}

\author[vicomtech]{Elena Zotova}
\ead{ezotova@vicomtech.org}
\address[vicomtech]{SNLT group at Vicomtech Foundation, Basque Research and Technology Alliance (BRTA)}

\author[ixaaddress]{Rodrigo Agerri\corref{mycorrespondingauthor}}
\ead{rodrigo.agerri@ehu.eus}
\cortext[mycorrespondingauthor]{Corresponding author}

\author[ixaaddress]{German Rigau}
\ead{german.rigau@ehu.eus}

\address[ixaaddress]{HiTZ Center - Ixa, University of the Basque Country UPV/EHU}

\begin{abstract}
Popular social media networks provide the perfect environment to study
the opinions and attitudes expressed by users. 
While interactions in social media such as Twitter occur in many natural languages, research on
stance detection (the position or attitude expressed with respect to a
specific topic) within the Natural Language Processing field has largely been done for English.
Although some efforts have recently been made to develop annotated data
in other languages, there is a telling lack of resources to
facilitate multilingual and crosslingual research on stance detection. This is
partially due to the fact that manually annotating a corpus of social media
texts is a difficult, slow and costly process. Furthermore, as stance is a
highly domain- and topic-specific phenomenon, the need for annotated data is
specially demanding. As a result, most of the manually labeled resources are hindered by
their relatively small size and skewed class distribution.
This paper presents a method to obtain multilingual datasets for stance
detection in Twitter. Instead of manually annotating on a per tweet basis, we
leverage user-based information to semi-automatically label large amounts of
tweets. Empirical monolingual and cross-lingual experimentation and qualitative analysis show
that our method helps to overcome the aforementioned difficulties to
build large, balanced and multilingual labeled corpora. We believe that our
method can be easily adapted to easily generate labeled social media data for other Natural
Language Processing tasks and domains. 
\end{abstract}


\end{frontmatter}



\section{Introduction}\label{sec:intro}
The phenomenon of \emph{fake news} is becoming notoriously common, particular
within social media. Fake news have been defined as ``a made-up story with an
intention to deceive''\footnote{\url{https://www.nytimes.com/2016/12/06/us/fake-news-partisan-republican-democrat.html}},
often for a secondary gain, and it is considered to be one of the most serious
challenges facing the news industry and the political sphere. 

Determining the veracity of a given article or social media message, in absence
of further context or background knowledge, is often a very difficult task,
even for expert fact-checkers. Thus, the organizers of the Fake News Challenge
considered that \emph{fake news} detection should be broken down into
intermediate tasks\footnote{\url{http://www.fakenewschallenge.org/}}, so that
the output of each of them would provide an \emph{indicator} to be taken into
account in the overall fake news detection task. The first stage of the Fake News
Challenge was stance detection. In their view, a stance detection system would
allow fact-checkers to automatically know which documents, messages or users
agree or disagree with a given document thereby helping to identify
contentious content. 

Stance detection has been addressed from at least two general perspectives, 
depending on whether the topic is open-ended or static. There are at least two
well known shared tasks which formulate open stance detection tasks. The
aforementioned Fake News Challenge, where the task is to classify whether a
given document \emph{agrees}, \emph{disagrees}, \emph{discusses} or is
\emph{unrelated} with respect to a previously published headline or news
document. Closely related to this, the RumourEval 2017 shared task
\citep{derczynski-etal-2017-semeval} referred to four different categories to
express the stance of a tweet with respect to a triggering message (often a
rumour), namely, \emph{support}, \emph{deny}, \emph{query} and \emph{comment}.
Closer to our work, the second perspective defines stance with respect to a
pre-defined or given topic or target, as it is often called. Thus, given a tweet and a
target entity or topic, Natural Language Processing (NLP) systems should try to
classify whether the stance expressed is in \emph{favour}, \emph{against}, or
whether is unrelated or \emph{none} with respect to the given target. Let us
consider the following two examples:

\begin{quote}
\textbf{Tweet:} \textit{I still remember the days when I prayed God for strength.. then suddenly God gave me difficulties to make me strong. Thank you God! \#SemST}

\textbf{Target:} Atheism

\textbf{Stance:} AGAINST

\vspace{0.5cm}

\textbf{Tweet:} \textit{@PH4NT4M @MarcusChoOo @CheyenneWYN women. The term is women. Misogynist! \#SemST}

\textbf{Target:} Feminist Movement

\textbf{Stance:} FAVOR
\end{quote} 

These two examples illustrate the nature of the task. Messages 
are very short, contain non-standard spelling
grammar, emojis, hashtags and figurative language such as irony and
sarcasm. This particular task, as defined by the Stance Detection in
Twitter at SemeEval 2016 \citep{mohammad-etal-2016-semeval}, consists of
classifying single tweets, without conversational structure, into one of three
classes: FAVOR, AGAINST and NONE. 

As it is often the case in the NLP field, research on stanced detection has mostly be done for
English, although recently there have been efforts to develop annotated corpora
for stance detection in languages other than English. For instance,
\cite{mohtarami-etal-2019-contrastive} experiments on the Arabic corpus provided by
\cite{baly-etal-2018-integrating}. A dataset in Czech was developed from
comments of news and used to experiment with SVM, Maximum Entropy and
Convolutional Neural Networks (CNNs) \citep{SLON2017-Stance}. Furthermore,
\cite{Vychegzhanin2019} presented a method of assembling classifiers for stance
detection in Russian. Finally, \cite{evrard-EtAl:2020:LREC} created
a French Twitter corpus for stance detection.

Still, there is a clear need for resources to investigate crosslingual
approaches to stance detection. To the best of our knowledge, the Catalan and
Spanish corpus provided by the IberEval 2017 and 2018 shared tasks
\citep{taule17,taule18} is the first work with the aim of facilitating
multilingual research on stance detection. This was later complemented by
\cite{Lai2020}, who provided another two datasets in French and Italian.
However, the datasets are about different topics on data obtained from
different dates which makes it very difficult to perform crosslingual research.
Moreover, the majority of the previous resources, both monolingual or
multilingual, are hindered by their small size and skewed class distribution
\citep{taule18,Lai2020}. This is partially due to the fact that manual
annotation on a tweet basis is a difficult, slow and costly task. To make
things worse, stance is a highly domain- and topic-specific phenomenon, which
means that each topic requires its own annotated dataset to develop
state-of-the-art classifiers. This creates an endless demand of labeled data. 

This paper tackles these issues by proposing a method to semi-automatically
obtain multilingual annotated data for stance detection based on a
categorization of Twitter users. The result of applying such method is the
\emph{Catalonia Independence Corpus} (CIC).

For our new corpus we collect and semi-automatically annotate tweets
(coetaneous, and on the same topic) in two different languages: Catalan and
Spanish. The availability of multilingual annotated
data collected on the same dates and on the same topics facilitates comparison
of models in crosslingual experimentation. Otherwise, it would be difficult
to know if differences in performance are due to the learning models or to
differences in topics and temporality of the obtained data. 

Our present work substantially improves and extends a first version of the CIC
dataset and the preliminary set of experiments presented in
\cite{zotova-etal-2020-multilingual}. More specifically, in this work we make
the following contributions. 

First, we devise an alternative method to build
a new version of the CIC dataset, CIC-Random, where the messages across the
train, development and test splits are of different users. 

Second, we provide a large set of experiments in four different datasets
(SemEval 2016 and IberEval 2018, CIC and CIC-Random) for three languages
(Catalan, English, Spanish) showing that systems behave consistently across
datasets and languages; this in turn suggests that our methodology to build
annotated datasets for multilingual and cross-lingual stance detection in
Twitter helps to alleviate the difficulties faced by previous manual-based
efforts \citep{mohammad-etal-2016-semeval, taule18}. 

Third, we use the newly created CIC dataset to perform cross-lingual
experimentation, the first of its kind for stance detection, with large
pre-trained multilingual language models such as multilingual BERT and
XLM-RoBERTa \citep{Devlin19,Conneau2019XLMRoberta}, comparing zero-shot
approaches with translation-based strategies. These experiments also help to
provide some insights about the multilingual behaviour of the transformers.

Fourth, we perform extensive error analysis to better understand the pros and
cons of our method with respect to manual annotation. It seems that while the
semi-automatic nature of our method introduces some noise in the annotations,
user-specific information provides the extra context required to better label
the individual tweets. 

Fifth, we use the CIC-Random version to show that systems obtaining better
results on the original CIC data, as opposed to results in previous benchmarks
\citep{mohammad-etal-2016-semeval,taule18}, were not due to the systems
overfitting on specific users' writing style. This result, together with the
manual inspection and annotation of a sample of the CIC corpus, suggests that
for stance detection in Twitter, its quality is as good as the one we obtain
manually. 

Finally, in order to facilitate reproducibility of results we publicly
distribute the CIC and CIC-Random corpora and the variously pre-processed
versions of every corpus used in this
paper\footnote{\url{https://github.com/ZotovaElena/Multilingual-Stance-Detection}}.

The rest of the paper is organized as follows: Section \ref{sec:related-work}
describes previous approaches to stance detection. In Section
\ref{sec:datasets} we describe our methodology to build datasets for stance
detection, method that has been first employed to generate our new CIC corpus. The experimental setup is specified in Section
\ref{sec:experimental-setup}. Section \ref{sec:results} reports on our
monolingual and cross-lingual experiments, while Section \ref{sec:analysis}
provides an error analysis of the obtained results. We finish with some
concluding remarks in Section \ref{sec:conclusion}.

\section{Related Work}\label{sec:related-work}
The growing interest on stance detection is demonstrated by at least three
recent surveys addressing the topic. The first one revised opinion mining
methods in general with a special focus on stance towards products
\citep{wang2019survey}. Another recent survey study detailed research
work that modeled stance detection as a text entailment task
\citep{kuccuk2020stance}. This particular survey provided a broad coverage of
stance detection methods, including works from various research domains such as
Natural Language Processing (NLP), Computational Social Science, and Web science.
Furthermore, it also surveyed the modeling of stance using text, network-based, and behavioural
features. More recently, \cite{aldayel2020stance} covered new research
directions on stance detection in social media.

Automatic stance detection in social media is divided into two main approaches.
First, those that rely on \emph{traditional} machine learning models combining
hand-engineered features \citep{Mohammad:2017:SST:3106680.3003433} or static
word representations \citep{bohler-etal-2016-idi}. Second, those that are based
on the application of deep learning and neural networks
\citep{augenstein-etal-2016-stance,zarrella-marsh-2016-mitre,wei-etal-2016-pkudblab,
igarashi-etal-2016-tohoku}.

Although other well-known English datasets for stance detection
exist\footnote{\url{http://www.fakenewschallenge.org/}}
\citep{derczynski-etal-2017-semeval}, closer to our particular interest is the
dataset from the SemEval 2016 shared task for Stance Detection in Twitter
\citep{mohammad-etal-2016-semeval}. On the supervised setting of SemEval 2016,
\citep{Mohammad:2017:SST:3106680.3003433} obtained the best results using a SVM
classifier based on word n-grams and character n-grams features, outperforming
other deep learning approaches
\citep{zarrella-marsh-2016-mitre,wei-etal-2016-pkudblab}.  Later,
\cite{Aldayel2019} explored users interactions in Twitter and compared various
features, including on-topic content, network interactions, user's preferences,
and online network connections.


More recently, some other deep learning approaches improved over the SemEval 2016
official state-of-the-art results. For instance, \cite{du2017stance} proposed a neural network-based model
to incorporate target-specific information by means of an attention mechanism.
In the work of \cite{benton-dredze-2018-using} recurrent neural networks are
combined with pre-trained ``user embeddings'', which enrich the training data
with additional information in user-based level. \cite{sun-etal-2018-stance}
presented an hierarchical attention network to weigh the importance of various
linguistic information, and learn the mutual attention between the document and
the linguistic information. One more approach \citep{Wei2018} with end-to-end
deep neural model leverages attention mechanism to detect stance through target
and tweet interactions. The work of \cite{siddiqua-etal-2019-tweet} proposes a
neural ensemble model that combines two densely connected BiLSTMs, nested
LSTMs where  each  module is coupled with an attention mechanism.

Current best results on the SemEval 2016 dataset are reported by
\cite{ghosh2019stance}. They offer a systematic comparison of seven stance
detection methods and fine-tuned a masked language pre-trained model (BERT Large)
\citep{Devlin19} to report current state-of-the-art results on this particular
benchmark.

Neural network approaches have also been successful for
the SemEval 2016 Task B (weakly-supervised setting). For instance, apart from
the previously mentioned systems, \cite{augenstein-etal-2016-stance} proposed a
bidirectional Long-Short Term Memory (LSTM) encoding model. First, the target
is encoded by a LSTM network and then a second LSTM is used to encode the tweet
using the encoding of the target as its initial state.

Semi-supervised methods include building stance detection models with small
training sets. \cite{misra-etal-2016-nlds} proposes a data
augmentation method for small annotated datasets based on
stance-bearing hashtags. \cite{FraisierProximity2018} proposed an ensemble of
systems to model the stance detection task at user level complemented by
content--, interaction-- and geographic--based proximity of social network
profiles. 


Multilingual approaches based on the TW-10 corpus were developed for the ``MultiModal Stance Detection in tweets on Catalan \#1Oct
Referendum'' task at IberEval 2018. The best Spanish system
\citep{Segura-Bedmar18} consisted of a linear classifier with TF-IDF
vectorization, obtaining a final 28.02 F1 macro
score in the Spanish test data. The best result for Catalan
\citep{Cuquerella2018CriCaTM} consisted of
combining the Spanish and Catalan training sets to create a larger and more balanced
corpus. They experimented with stemming of various lengths (three, four and
five characters) and removing character suffixes from the word, which helped to
generalize over Catalan and Spanish. Their final F1 macro was 30.68.

\cite{Lai2020} propose a multilingual stance detection system for English,
Spanish, Catalan, French and Italian. The English, Catalan and Spanish data are
based on the SemEval 2016 and IberEval 2017 \citep{Lai2017iTACOSAI} corpus respectively, whereas the French and
Italian datasets were originally presented for that paper. They explore different
types of features---stylistic, structural, contextual, and affective---and
their contribution in the learning process of models such as BiLSTM, CNN
and SVM. Their scores for English, Catalan and Spanish were substantially
improved by \cite{zotova-etal-2020-multilingual}, a preliminary version of the
approach we present in this paper.


Summarizing, the few multilingual approaches presented so far are hindered by
the small size and skewed class distribution of the existing datasets. In the
next section we will
examine three of these resources, SemEval 2016 and TW-10 (Catalan and Spanish)
and propose our semi-automatic method to efficiently obtain good
quality annotated data for multilingual stance detection in Twitter.


\section{New Dataset: The Catalonia Independence Corpus}\label{sec:datasets}
In this section we provide a detailed description of the method
developed to generate the Catalonia Independence Corpus (CIC). 
In order to study multilingual and crosslingual approaches for stance detection in
Twitter, it is desirable to obtain annotated datasets on a common topic for more than one language and
obtained on the same dates (coetaneous). Previous works
include datasets in several languages, notably, IberEval 2018 \citep{taule18}
and \citep{Lai2020}. However, they do not provide an adequate setting for multilingual and
crosslingual studies to stance detection. 

With respect to IberEval 2018, they provide annotated data in Catalan and
Spanish, but in the Catalan part the classes distribution is extremely skewed
\citep{taule18}. Regarding \cite{Lai2020}, they developed two new datasets for French and Italian, but they
are not coetaneous nor about the same topic. These issues make crosslingual
experimentation very difficult as differences between performance across
languages may be due to other issues rather than model performance, namely, one
topic being more difficult than the other.

Finally, most previous datasets are quite small because they exclusively rely on
manual annotation on a per tweet basis. This is very costly but also not very
efficient because in many cases annotating a tweet without its background context
is nigh on impossible.

Our new dataset for stance detection in Twitter aims to address these
shortcomings: (i) the collected data is coetaneous across languages, (ii) on
the same topic, (iii) multilingual (Spanish and Catalan), (iv) their class
distribution is balanced, and (v), their annotation method more efficient and, as a
consequence, their size is much larger than previous datasets for stance
detection.

The first three issues are addressed in the data collection phase, but,
crucially, issues (iv) and (v) are direct consequences of our methodology to
obtain stance annotations. Our method for efficient annotation is based on the
following \emph{four building steps}:

\begin{itemize}
	\item \textbf{User-based annotation}: namely, taking into account the full
		timeline. We assume that it is easier to annotate a full timeline rather than the text of a single tweet without context.
	\item \textbf{User relations}: Based on previous research on political
	homophily \citep{barbera2015birds,himelboim2013birds,zubiaga2019political}, we use the relations between users
		(retweets) to obtain more users or accounts from which to obtain the
		tweets for our dataset.
	\item \textbf{Hashtags and Keywords}: A selection of hashtags and keywords
		are applied to obtain tweets that are relevant to our topic of
		interest. 
	\item \textbf{Topic Modelling}: We refine the extraction of on-topic tweets
		performed on step 2 by applying LDA. The idea is to cross user-level
		and topic information to provide annotated tweets for the final version
		of our dataset. 
\end{itemize}

The result is a final balanced dataset containing 10K annotated tweets for Spanish and
for Catalan, respectively. This new dataset is the largest annotated corpus of its kind, displaying all the
required features to perform crosslingual experimentation. In the following we
describe each of the steps involved in the development of the CIC corpus.

It should be noted that our annotation process, described in Sections
\ref{sec:categ-at-user}, \ref{sec:user-relations},
\ref{sec:hashtags} and \ref{sec:topic-detection}, is language- and
topic-independent, namely, the four steps listed above can be performed
regardless of the topic, target language and even task. However, these issues do affect the
collection of the data, as the objective of data collection must be obtaining
on-topic tweets that can then be leveraged for classification.

\subsection{Data Collection}\label{sec:data-collection}

In order to create the CIC dataset, we used a collection of tweets gathered
during 12 days on February and March 2019 in Barcelona and
Terrassa\footnote{An industrial city 25km away from Barcelona}. This collection
was originally obtained for industrial research in stance detection and
political ideology (left-right) prediction. The crawling was performed with
full access to the Twitter API. For its use in academic research, we first
separated them by
language\footnote{\url{https://code.google.com/archive/p/language-detection/}}
obtaining 680,000 tweets in Catalan and 2 million tweets in Spanish. We
discarded duplicated messages and those shorter than three words.

For the data collection, we first compiled a list of Twitter accounts from
media, political parties and political activists that clearly and explicitly
express their stance with respect to the independence of Catalonia. This list
was manually compiled based on the high visibility of users related to the political
scene in Catalonia. In total we obtain around 150 accounts of personalities,
political parties and digital media. 

Secondly, we extracted the most active and retweeted tweets in the crawled
corpus selecting a list of 1200 accounts in which there was enough content related
to the topic of interest, namely, the independence of Catalonia. In this step non-political accounts were
discarded. 

With respect to the data collection stage, it should be noted that in order to
have some data available to learn stance about a given topic we need tweets
that actually talk about that topic. That is true for any topic, not just for
the one in our dataset. Thus, if we were to collect data about other topics
such as \emph{Feminism} or \emph{Donald Trump} or \emph{Climate Change} or
\emph{Brexit}, we would still need to identify keywords, accounts and/or hashtags that talk about those topics.
Therefore, this is not an issue specific to our approach. 

\subsection{User-based Annotation}\label{sec:categ-at-user}

Annotation of the 1200 accounts obtained in the previous step 
was carried out using the same three labels and guidelines as in
previously existing stance datasets \cite{mohammad-etal-2016-semeval,taule18}.
Thus, FAVOR and AGAINST refer to a positive or negative stance towards the
independence of Catalonia, respectively. Finally, NONE will express neither a
negative nor a positive stance, or simply that it is not possible to reach a
clear decision.

Unlike previous approaches, and in order to increase the consistency of the
annotations and to speed up the annotation process, we do not manually annotate
each tweet in a one-by-one fashion. Instead, the annotation process was mainly based on
classifying stance at user level, namely, we categorized the tweets' authors manually by
checking their Twitter accounts. 

The assumption was that for a human annotator it is easier to annotate a full timeline
rather than the text of a single tweet without additional context. Thus, in our
annotation process the decision
about stance was also made taking into consideration other aspects from the
users' accounts, such as the use of special emojis and symbols that may state
clearly the stance towards the target (e.g., displaying a yellow ribbon is a
pro-independence symbol, whereas a Spanish flag would convey that the user is
against the independence, etc.), or by the Bio section. In this step each user
or Twitter account is assigned a stance label: FAVOR, AGAINST or NONE.

Table \ref{tab:example_against} presents an example of a user categorised as
being AGAINST the independence of Catalonia. These tweets were written by
Manuel Valls, a Catalan politician from a unionist party. He clearly
expresses his opinion about the topic with typical vocabulary such es {\it coup
d'état}, {\it separatist} or {\it constitutionality}. 


\begin{table}[]
\resizebox{\textwidth}{!}{%
\begin{tabular}{p{8cm}p{8cm}} 
\hline
\textbf{Tweet by @manuelvalls} & \textbf{Translation} \\ \hline
Pese a la complicidad de Chavists del Ayuntamiento de Barcelona  como los populistas de izquierdas y los independentistas desde ?@CiudadanosCs?  y ?@CiutadansBCN? llevamos años luchando por la libertad y el respeto  a los derechos humanos. \#VenezuelaEnLaCalle \#Venezuela https://t.co/fjw1hxXyYD\textbackslash{}end & Despite the complicity of Chavistas from the Barcelona  City Council such as the left-wing populists and the  independentists from ?@CiudadanosCs? and ?@CiutadansBCN?  we have been fighting for freedom and respect for human rights  for years. \#VenezuelaEnLaCalle \#Venezuela https://t.co/fjw1hxXyYD \\
& \\
El relato del golpe de Estado en Cataluña contado por los protagonistas constitucionalistas. Vargas Llosa: ``Era algo gigantesco. Creo que la manifestación más grande que he visto''. Por @e\_bece  https://t.co/Aymvhgj6cw & The story of the coup d'état in Catalonia told by the constitutionalist protagonists. Vargas Llosa:``It was  something gigantic. I think it was the biggest demonstration  I've ever seen''. By @e\_bece https://t.co/Aymvhgj6cw \\
 & \\
La noche que el Rey defendió a España: ``El 3 de octubre de 2017 el Rey cortó en seco una ensoñación separatista que llevaba años marcando goles fuera de juego al Estado ante unos  árbitros que se limitaban a llevarse las manos a la cabeza''. Por @AlmudenaMF https://t.co/XDDM2D96XH & The night the King defended Spain: ``On 3 October 2017, the King cut short a separatist dream that had been scoring offside goals for the State for years in front of referees who were simply putting their hands on their heads.'' By @AlmudenaMF https://t.co/XDDM2D96XH \\ \hline
\end{tabular}}
\caption{Example of a user (@manuelvalls) manually classified as AGAINST of independence of Catalonia.}
\label{tab:example_against}
\end{table}

\begin{figure}[!ht]
\centering
\includegraphics[width=0.4\textwidth]{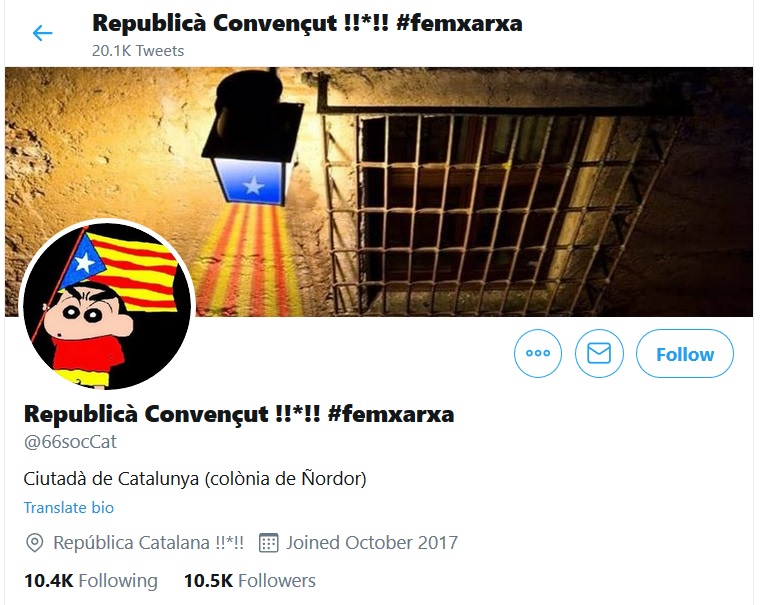}
\includegraphics[width=0.4\textwidth]{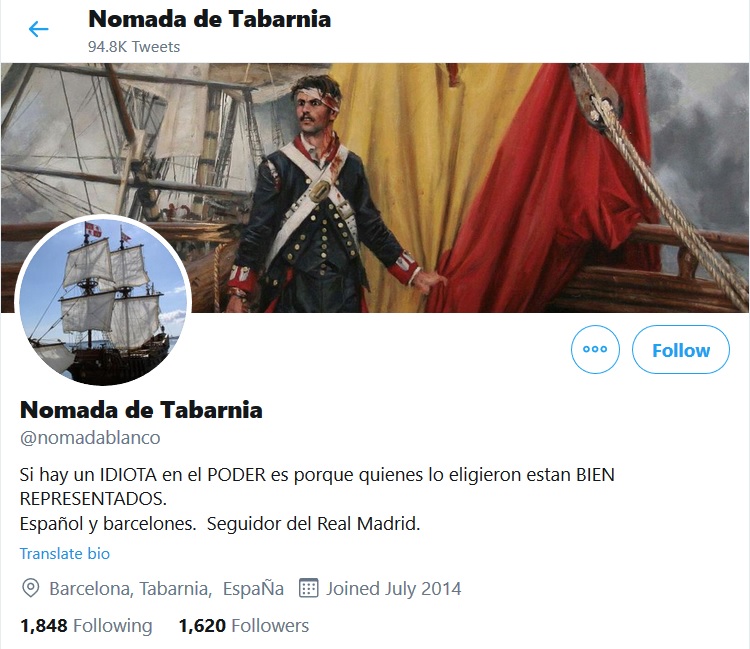}
\caption{Two examples of accounts which can be easily annotated as FAVOR and
AGAINST, respectively. The Catalonian republican flag appears on the FAVOR-user's profile whereas the Spanish official flag is displayed in the profile of the user AGAINST the independence.}
\label{fig:twitter_account}
\end{figure}

Furthermore, Figure \ref{fig:twitter_account} presents two {\it simple}
examples of accounts which can be easily categorized as FAVOR (left) and
AGAINST (right), just by looking at their profile. Thus, the FAVOR-user displays a
flag related to the Catalonian Republic and the independence movement.
Moreover, the user against independence displays a profile related to Tabarnia, an unionist symbol\footnote{\url{https://en.wikipedia.org/wiki/Tabarnia}}.

\subsection{User Relations}\label{sec:user-relations}

In addition to the user-level annotations, we extracted the relations between users based on their retweets
\citep{SNA2002}. A study of the behaviour of Twitter users \citep{boyd_retweets}
mentions, among others, that retweets are motivated by ``publicly agreeing
with someone'', ``to give visibility'', or ``to validate other's thoughts''.

Furthermore, previous research on political homophily
\citep{barbera2015birds,himelboim2013birds} shows that homophily is also reflected in social
media. In other words, supporters of one particular party or ideology are more
likely to interact with users of the same ideology or party.

This idea has been studied in several NLP works relevant to this work
\citep{Lai2020,zubiaga2019political}. In particular, \cite{zubiaga2019political} empirically demonstrates
that there is a correlation between the national identity of users (in relation
to independentist movements) and the relations between users in the social
network. Their case study also includes Catalonia, together with Scotland and
the Basque Country. For our particular interests, this means that by using the retweets, we can increase the size of our initial
data pool (1200 users) with more users that are anti- and pro-independence.

Following the principle of political homophily, we assumed
that in general users retweet mostly users expressing the same political stance as the original
message. While this method may introduce some noise, it allowed us to quickly
obtain a large amount of annotated data. Thus, from the 1,200 accounts that were
labelled manually in the previous step, we were able to automatically obtain 25,510
categorized users. We do not distinguish between Catalan and Spanish users
because most of them are fully bilingual. Table
\ref{table:users} reports the distribution of the categorized users. The final
set contains 131,022 unique tweets in Catalan and 202,645 unique tweets in
Spanish.

\begin{table}[!ht]
\centering
\begin{tabular}{lr} \hline
      \textbf{Label} & \textbf{Count} \\ \hline
      AGAINST & 3,091 \\
      FAVOR & 22,247 \\
      NONE & 176 \\ \hline
\end{tabular}
\caption{Distribution of the categorized users.}\label{table:users}
\end{table}

Thus, while the use of user relations (retweets) is technically motivated by
the need to increase the number of accounts from which to extract the tweets to
be included in the final dataset, this is theoretically and empirically
motivated by previous research on political homophily
\citep{barbera2015birds,himelboim2013birds}, and especially on independence
movements \citep{zubiaga2019political}.

\subsection{Hastags and Keywords}\label{sec:hashtags}

As we mentioned earlier, we annotated every tweet in the corpus by assigning
the stance directly to the account users. However,
this does not mean that we can use every tweet from the users, given that many
messages may not be related to our specific target, namely, the independence of
Catalonia. This issue is addressed by extracting relevant tweets using hashtags
and keywords and by applying LDA topic modelling. The latter is described in
the next section. 

In this step we extracted all the hashtags from the corpus
and selected manually those that were related to the independence of Catalonia,
such as \textit{\#Catalu\~naesEspa\~na, \#CatalanRepublic, \#Tabarnia,
\#GolpeDeEstado, \#independ\'encia, \#judicifarsa, \#CatalanReferendum} etc.,
totalling 450 hashtags. We also manually added keywords for both languages, 25 in total.
We labeled each tweet as being on topic if it contained one of the relevant
hashtags or keywords. Table \ref{tab:relevanttweets} displays the distribution
of tweets after applying this filtering step using hashtags and keywords.

\begin{table}[!ht]
\centering
\begin{tabular}{lrr} \hline
\textbf{Label} & \textbf{Catalan} & \textbf{Spanish}\\ \hline
AGAINST & 1,476 & 8,267 \\
FAVOR & 23,030 & 11,843 \\
NONE & 986 & 497 \\ \hline
\end{tabular}
\caption{Distribution of tweets obtained by hashtags and keywords related to
the ``Independence of Catalonia''.}\label{tab:relevanttweets}
\end{table}

\subsection{Topic Detection}\label{sec:topic-detection}

The distribution of tweets per language obtained in the previous step, as shown
by Table \ref{tab:relevanttweets}, evidences that the
vast majority of the tweets are labelled as FAVOR. As our aim was to obtain a
balanced dataset, we needed to add more tweets to
the under-represented classes. We use the MALLET \citep{McCallumMALLET}
implementation of Latent Dirichlet allocation (LDA)
\citep{Blei:2003:LDA:944919.944937} to perform basic target detection
in the corpus of categorized users described in Table
\ref{table:users}, Section \ref{sec:user-relations}. The objective was to obtain more on-topic tweets for those
classes that are under-populated (AGAINST and NONE in Catalan and NONE in Spanish). We
manually revised the obtained topics and selected only those tweets which were
clustered within the \emph{independence} topic.



In a final step, we selected approximately 10,000 tweets per language
(excluding those shorter than four words) keeping the proportion of users from
the initial pool of crawled tweets. We split them keeping 60\% for training,
and 20\% each for development and test. The average length of a tweet in the
Catalan Independence Corpus is very similar to the average in the TW-1O
dataset. Considering that our corpus does not include extra context, this means
that our average length is longer than those obtained at SemEval 2016 or TW-1O.
Futhermore, our corpus is also much larger
\citep{mohammad-etal-2016-semeval,taule18} and presents a more balanced
distribution of classes, as shown by Table \ref{tab:cic_distribution}. The
result of this final step is the \emph{Catalonia Independence Corpus} (CIC).










\subsection{User Bias} \label{subsection:bias}

A common feature of corpora based on social media is that a small amount of
users usually generate a large proportion of the posts and viceversa. We
checked this issue in the original CIC corpus and found out that in the Spanish
subset there are 407 users whereas for Catalan the authors were 1,100.
Furthermore, we also realized that the same users were present across the three partitions, train, development
and test. As tweets from the same author might contain specific information about
the user, such as individual writing style, vocabulary or other communicative
behaviour, we wanted to double check that systems were not learning those
specific features and thus overfitting to the characteristics of some specific
users.


For that reason, we created a new version of the original CIC dataset,
namely, CIC-Random, reorganizing the CIC corpus in such a way that
users appearing in the train set were not included in the development and test
sets. More specifically, we randomly sampled
the list of users and created three new splits for training, development
and test, according to three criteria: (i) tweets from the same user cannot
occur across the three splits; (ii) the proportion of tweets of each user
in each split has to be same as in the original CIC corpus and, (iii) the
balance between the three stance labels must be kept. The result of this process is
the \emph{CIC-Random} dataset.

As randomness was the key in this process, the distribution is
not exactly the same as in the original CIC corpus, as shown by Table
\ref{tab:cic_distribution}. 

\begin{table}[]
\centering
\small{
\begin{tabular}{lrrrrrrrrrr} \hline
\textbf{} & \multicolumn{3}{r}{\textbf{Train}} & \multicolumn{3}{c}{\textbf{Validation}} & \multicolumn{3}{c}{\textbf{Test}} & \multicolumn{1}{c}{\textbf{Total}} \\ \hline
\textbf{Dataset} & \textbf{A} & \textbf{F} & \textbf{N} & \textbf{A} & \textbf{F} & \textbf{N} & \textbf{A} & \textbf{F} & \textbf{N} & \textbf{} \\ \hline
CIC-CA & 2,680 & 2,545 & 752 & 1,201 & 506 & 372 & 937 & 850 & 205 & 10,048 \\
CIC-CA-Random & 2,416 & 2,335 & 1,277 & 820 & 763 & 427 & 804 & 752 & 454 & 10,048 \\
CIC-ES & 2,560 & 2,276 & 1,100 & 875 & 702 & 503 & 953 & 843 & 265 & 10,077 \\
CIC-ES-Random & 2,515 & 2,420 & 1,111 & 856 & 782 & 377 & 829 & 807 & 380 & 10,077 \\ \hline
\end{tabular}
\caption{Distribution of the CIC corpus separated into training, validation and test sets.  A---AGAINST, F---FAVOR, N---NONE. }
\label{tab:cic_distribution}
}
\end{table}

\section{Experimental setup}\label{sec:experimental-setup}
As it is explained in the previous chapter, the development of the Catalonia 
Independence Corpus (CIC) was motivated by the lack of large, balanced,
multilingual and coetanous corpora for stance detection. Indeed, previous experiments have
shown the difficulty of drawing meaningful conclusions with the TW-1O corpus,
mainly due to the skewed class distribution in the Catalan set
\citep{taule18,zotova-etal-2020-multilingual}.

In this section we describe the setup for the experiments performed using the
dataset described in the previous chapter plus two previously existing
benchmarks, SemEval 2016 and TW-10. Our motivation is to show that, in addition
to being faster and cheaper to generate Twitter-based annotated datasets for
stance detection, our semi-automatic, user-based method produces annotated data
which is as reliable for experimentation as manually-based annotated datasets.
We will investigate this claim by studying the behaviour of popular text
classification baselines and methods across datasets, both monolingual (SemEval
2016) and multilingual (TW-1O and CIC). We would expect the systems to exhibit
similar behaviour across datasets. 

Moreover, the development of a multilingual dataset such as CIC
allows us to experiment with the application of
large multilingual language models such as mBERT or XLM-RoBERTa
\citep{Devlin19,Conneau2019XLMRoberta} in cross-lingual settings. Thus,
for a scenario in which there is not training data available for the target
language, we can investigate whether it would be better: (i) using the multilingual
language model for the target language directly in a zero-shot fashion or, (ii)
automatically translating the data from the target language to a language
for which we do have training data available, namely, applying a \emph{``translate and fine-tune''} method.

In the rest of this section we describe the benchmark dataset, the data-preprocessing methods and the systems
used for experimentation. We used four different system types: (i) 
TF-IDF vectorization with a SVM classifier (TF-IDF+SVM);
(ii) SVM trained by averaging fastText word embeddings \citep{Grave18} for the
representation of tweets (FTEmb+SVM); (iii) the fastText text classification
system \citep{joulin-etal-2017-bag} with fastText word embeddings
(FTEmb+fastText), (iv) pre-trained language models based on the transformer
architecture, including monolingual (XLNET \citep{Yang2019XLNetGA} and RoBERTa
\citep{Liu2019RoBERTaAR}) and multilingual language models (mBERT \citep{Devlin19} and
XLM-RoBERTa \citep{Conneau2019XLMRoberta}). 

\subsection{Benchmark Datasets}\label{sec:benchmark-datasets}

In this section we describe other two well-known datasets that, together with
our newly created CIC Corpus, will be used for the experimentation. This means that our experiments will be
evaluated on seven different datasets and three
languages, namely, Catalan English, and Spanish. We first described 
the SemEval 2016 dataset on Stance Detection, the first of its kind, and then
the multilingual TW-10 corpus. 

\subsubsection{SemEval 2016}\label{sec:semeval2016}

The dataset presented at the Stance Detection task organized at SemEval
2016\footnote{\url{http://alt.qcri.org/semeval2016/task6/data/}}
\citep{mohammad-etal-2016-semeval}, consists of English tweets labeled for
both stance (AGAINST, FAVOR and NONE). In the supervised
track, more than 4,000 tweets are annotated with respect to five targets:
``Atheism'', ``Climate Change is a Real Concern'', ``Feminist Movement'',
``Hillary Clinton'', and ``Legalization of Abortion''. For each target, the
annotated tweets were ordered by their timestamps. The first 70 percent of the
tweets form the training set and the last 30 percent are reserved for the test
set. Table \ref{tab:semeval_dataset} presents the final distribution of
training and testing examples.

\begin{table}[h]
\centering
\begin{tabular}{lcc}
\hline
\textbf{Target}                  & \textbf{Train} & \textbf{Test} \\ \hline
Atheism                          & 513            & 220           \\ 
Climate Change is a Real Concern & 395            & 169           \\ 
Feminist Movement                & 664            & 285           \\ 
Hillary Clinton                  & 639            & 295           \\ 
Legalization of Abortion         & 603            & 280           \\ \hline
\textbf{Total}                   & \textbf{2,814}  & \textbf{1,249} \\ \hline
\end{tabular}
\caption{Number of examples per target in the SemEval 2016 English dataset.}
\label{tab:semeval_dataset}
\end{table}

To prepare the dataset the organizers collected 2 million tweets containing
FAVOR, AGAINST and ambiguous (NONE) hashtags for the selected targets. These
hashtags were removed after manual annotation, which was performed via
crowdsourcing by eight different annotators. In
addition to stance, annotations are provided to express whether the target is
explicitly mentioned in the tweet. Table \ref{tab:semeval_dataset} shows some
examples taken from the SemEval 2016 dataset.

\begin{table}[!ht]
\resizebox{\textwidth}{!}{%
\begin{tabular}{p{8cm}p{2cm}p{2cm}} \hline
\textbf{Tweet} & \textbf{Target} & \textbf{Stance} \\ \hline
\textit{@PH4NT4M @MarcusChoOo @CheyenneWYN women. The term is women. Misogynist! \#SemST} &  Feminist Movement &  FAVOR \\ 
 & & \\ 
\textit{American conservatism has everything to do with religion with all the good stuff taking out of it. \#SemST} & Atheism & AGAINST \\ \hline
\end{tabular}}
\caption{Examples of tweets from SemEval 2016 dataset.}
\label{tab:tweet_semeval}
\end{table}

The best result was reported by the organizers of the task with a system based
on character and word n-grams to train a linear SVM model, achieving 68.98\% in
F1 average score \citep{mohammad-etal-2016-semeval}. More recently,
\cite{ghosh2019stance} fine-tuned the pre-trained BERT Large model
\citep{Devlin19} obtaining the highest performance so far on this dataset, with
a F1 average score of 75.1.



\subsubsection{TW-1O Referendum}\label{sec:referendum}

The ``MultiModal Stance Detection in tweets on Catalan \#1Oct
Referendum'' shared task at IberEval 2018
(MultiStanceCat) proposed to detect stance (FAVOR, AGAINST, NEUTRAL) on political discourse with respect to
the Referendum on the Independence of Catalonia held on the first of October,
2017. The dataset is multilingual (Catalan and Spanish) and includes images to
facilitate multimodal experimentation \citep{taule18}. 

The dataset was collected using the hashtags \#1oct, \#1O, \#1oct2017 and
\#1octl6 to search for messages in Twitter, widely used in the dates previous
to the Referendum.  A total of 87,449 tweets in Catalan and 132,699 tweets in Spanish were collected
between between September 20-30, 2017. After various pre-processing steps, the final dataset consists of 11,398
tweets: 5,853 written in Catalan (the TW-1O-CA) and 5,545 in Spanish
(the TW-1O-ES). The dataset was annotated manually by three experts.
Also, the previous and next messages are included as additional context to the original tweet. 
Contatenating the three tweets
result in an approximate average length of 38 tokens per document. 







Table \ref{tab:twdatasetdistr} illustrates the classes distribution in the
TW-10 data. While for Spanish the distribution of classes is quite balanced, in
Catalan the FAVOR class occurs 35 times more than AGAINST, and 8 times more
than NEUTRAL. Obviously, this hugely skewed dataset would make it difficult to build and compare models
for Catalan and across languages. In this sense, a simple most frequent class
baseline for Catalan would be correct in around 95\% of the cases.

\begin{table}[!ht]
\centering
\begin{tabular}{lrr} \hline
      \textbf{Label} & \textbf{Catalan} & \textbf{Spanish}\\ \hline
    AGAINST & 120 & 1,785 \\
      FAVOR & 4,085 & 1,680 \\
     NEUTRAL & 479 & 972 \\ \hline
      Total & 4,684 & 4,437 \\ \hline
\end{tabular}
\caption{Distribution of classes in the TW-1O trainset.}\label{tab:twdatasetdistr}
\end{table}


\subsection{Data Pre-processing}\label{subsec:data-pre-processing}

Each of the system types mentioned above benefit from different pre-processing
strategies. We believe this is particularly important as the type of
pre-processing has a huge influence in the final performance. We followed four
different pre-processing strategies, illustrated by Table
\ref{tab:text_preproc}.

\begin{table}[]
\centering
\resizebox{\textwidth}{!}{%
\begin{tabular}{p{2cm}p{12cm}}  \hline
\textbf{Pre-processing type} & \textbf{Result}     \\ \hline
Original   & @pilarc\_pilarc Ten, manipuladora te cayó el ME ESTAS HABLANDO EN POLACO?? que le suelta el fachamierda primero \#ZASCA https://t.co/XQ08KuVgtI \\ \hline
Type A  & manipulador cayo hablar polaco suelto fachamierda \#zasca \\ \hline   
Type B   & ten manipuladora se te cayó el me estas hablando en polaco que le suelta el fachamierda primero \#zasca \\ \hline
Type C  & pilarcpilarc Ten manipuladora se te cayó el ME ESTAS HABLANDO EN POLACO  que le suelta el fachamierda primero ZASCA  \\ \hline
Type D   & pilarc\_pilarc Ten, manipuladora se te cayó el ME ESTAS HABLANDO EN POLACO??  que le suelta el fachamierda primero ZASCA \\ \hline  
\end{tabular}}
\caption{Examples of the four types of text pre-preprocessing.}
\label{tab:text_preproc}
\end{table}

\noindent \textbf{Type A}: Punctuation, URLs, retweets (RTs), Twitter
usernames, hashtags, digits, stopwords, words shorter than three characters,
diacritics and emojis are removed. Furthermore, multiple character repetition
is replaced with a single chararacter. The text is lemmatized and lowercased.
Lemmatization is performed via dictionary look-up\footnote{\url{https://github.com/michmech/lemmatization-lists}}. Note that
this lemmatization method does not deal with ambiguity.

\noindent \textbf{Type B}: We remove punctuation, URLs, RTs, digits, Twitter
usernames, hashtags and emojis. Repeated characters are simplified into a
single character. Text is lemmatized and lowercased.

\noindent \textbf{Type C}: Punctuation, Twitter usernames, hashtags and URLs are removed. 

\noindent \textbf{Type D}: Minimal pre-processing: Twitter usernames, hashtags
and URLs are removed. 

\subsection{TF-IDF+SVM}\label{sec:tf-idf+svm}

TF-IDF (Term Frequency times Inverse Document Frequency)
\citep{Jones72astatistical} is a weighting scheme broadly used in many tasks.
Its goal is to reduce the impact of words that occur too frequently in a given
corpus. TF-IDF is the product of two metrics, the term frequency and the
inverse document frequency. We calculate the TF-IDF scores for all
pre-processed unigrams in the training corpus. The number of features equals
the size of the vocabulary of the dataset and represents the dimensionality of
the document vector. The TF-IDF vectorizing is applied over the text
pre-processed following Type A strategy. 

We also used Information Gain \citep{Cover:2006:EIT:1146355} for feature
selection and Grid search for hyperparameter optimization. The Information Gain scores show how
common a specific feature is in a target class. For example, those words that
occur mainly in tweets labelled as FAVOR will be highly ranked. All the weights
are normalized and the features ranked from one to zero. We then select those
features that are larger than zero. Grid search is used to tune
two SVM (RBF kernel) hyperparameters, namely, C and gamma.

%

\subsection{FastText Embeddings+SVM}\label{sec:svm+f-embedd}

Word embeddings encode words as continuous real-valued representations in a low
dimensional space. Word embedding models are pre-trained over large corpora and are able
to capture semantic and syntactic similarities based on co-ocurrences.

FastText distributes static word embeddings for our languages of interest,
namely, Catalan, English and Spanish \citep{Grave18}.
Initial experimentation showed that the Common
Crawl\footnote{\url{http://commoncrawl.org/}} models performed better for the
stance detection task. The Common Crawl models are trained using a Continuous
Bag-of-Words (CBOW) architecture with position-weights and 300 dimensions on a
vocabulary of 2M words. In order to produce vectors for out-of-vocabulary
words, fastText word embeddings are trained with character n-grams of length 5,
and a window of size 5 and 10 negatives \citep{Grave18}. We represent the tweet
as the average of its word vectors \citep{DBLP:journals/corr/KenterBR16}. In
order to facilitate the look-up into the pre-trained word embedding model, we
use the Type B pre-processing strategy. As for the previous system, C and
gamma hyperpameters (SVM RBF kernel) are configured via grid search. 

\subsection{FastText System}\label{sec:fasttext-system}

Apart from the pre-trained word embedding models, fastText also refers to a
text classification system \citep{joulin-etal-2017-bag}. The fastText system
consists of a linear model with rank constraint. First a weight matrix A is
build via a look-up table over the words. Then the word representations are
averaged to construct the tweet representation, which is then fed into a linear
classifier. This is similar to the previous approach, but in the fastText
system the textual representation of the tweet is a hidden variable which can
be reused. The CBOW model proposed by \cite{mikolov2013b} is similar to this
architecture, with the difference that the middle word is replaced by the
stance label. Finally, fastText uses a softmax function to calculate the
probability distribution over the predefined classes.

\subsection{Transformers}\label{sec:neural-architecture}

BERT \citep{Devlin19} is a pre-trained mask language model based on the
transformer architecture \citep{VaswaniSPUJGKP17} which has obtained very good
results on many NLP tasks. The first multilingual version of such models was
multilingual BERT (mBERT), a single language model pre-trained from corpora in
more than 100 languages. Another well-known model is XLM-RoBERTa
\citep{conneau2019unsupervised}, which provides a language model for 100
languages trained on 2.5 TB of Common Crawl text. Both mBERT and XLM-RoBERTa
allow to perform cross-lingual knowledge transfer
\citep{heinzerling-strube-2019-sequence,DBLP:conf/acl/PiresSG19,karthikeyan2020cross},
namely, these systems can be applied to generate predictions on datasets for
languages different to the ones used to fine-tune them. Thus, in this paper we will use both mBERT and
XLM-RoBERTa for fine-tuning in Catalan and Spanish but also in cross-lingual
experiments using the CIC corpus. Additionally, we will use two other transformers which offer pre-trained
English language models: RoBERTa and XLNet.


RoBERTa \citep{Liu2019RoBERTaAR} is an improved, optimized version of
BERT. The model was trained using BERT architecture with larger batches and
much more data (around 160GB of English texts). Furthermore, the next sentence prediction objective used for pre-training the language model
is discarded. 

XLNet \citep{Yang2019XLNetGA} is an auto-regressive method based
on permutation language modelling \citep{uria_aotoregressive} without using any
masking symbols. XLNet integrates two-stream self-attention and a Transformer-XL architecture
\citep{Dai19} to the pre-training process with the objective of learning long-range
dependencies.

Every experiment with the transformers uses the datasets
pre-processed following the Type D strategy, which minimally removes hashtags, Twitter
usernames and URLs from the tweets. 

Furthermore, we use the base version of each transformer so that we can fine-tune the pre-trained model in a basic GPU of 12GB RAM.
For each dataset, we tune the hyperparameters (batch size, minimum sequence
length, learning rate and number of epochs) on the development data if
available, otherwise, the training set is used both for training and
development.

\subsection{Evaluation}\label{sec:evaluation}

The models are evaluated with the metric and script provided by the organizers of SemEval 2016
\citep{mohammad-etal-2016-semeval} which reports F1 macro-average score of two
classes: FAVOR and AGAINST, although the NONE class is also represented in the
test data:

\[F1_{avg} = \frac{F1_{favor} + F1_{against}}{2}\]

\clearpage
\section{Experimental Results}\label{sec:results}
In this section we report the results obtained by applying the
experimental setup from Section \ref{sec:experimental-setup} to the
datasets described in Section \ref{sec:datasets}.

More specifically, in this section we will show, via experimentation, that our
proposed method to develop the CIC corpus generates multilingual datasets for
stance detection in Twitter which are at least as reliable for experimentation
as manually annotated ones. In this sense, our method would help to facilitate
the development of new multilingual and cross-lingual approaches for stance
detection.

In the following, we first report the results obtained
for each of the datasets in a standard in-domain setting. Finally, we perform
cross-lingual experiments on the CIC corpus using multilingual language models
(mBERT and XLM-RoBERTa).

\subsection{SemEval 2016}\label{sec:results-semeval}

In Table \ref{tab:semeval-results} we can see the results of the experiments
performed on the SemEval 2016 dataset. The first three rows refer to the
systems based on linear classification, namely, SVM using TF-IDF
\ref{sec:tf-idf+svm}, SVM with averaged fastText embeddings
\ref{sec:svm+f-embedd}, and the fastText system itself
\ref{sec:fasttext-system}. The next four rows provide the results obtained
using the monolingual (XLNet and RoBERTa) and multilingual (mBERT and
XLM-RoBERTa) Transformer-based pre-trained language models.

\begin{table}[!ht]
\centering
\begin{tabular}{llll} \hline
\textbf{System} &  \textbf{F1$_{against}$} & \textbf{F1$_{favor}$} & \textbf{F1$_{avg}$}\\ \hline
TF-IDF+SVM & 73.24 & 53.52 & 63.38 \\
FTEmb+SVM & 72.06 & 53.56 & 62.81 \\
FTEmb+fastText & 72.94 & 61.58 & 67.26 \\
mBERT \small{(msl 256, batch 32, lr 5e-5, 5e)} & 73.90	& 62.29 & 68.10 \\
XLM-R \small{(msl 128, batch 16, lr 2e-5, 10e)} & 75.30	& 65.03	& 70.17 \\
XLNet \small{(msl 128, batch 16, lr 2e-5, 10e)} & 76.23	& 66.57	& 71.40 \\
RoBERTa \small{(msl 256, batch 16, lr 2e-5, 10e)} &	76.87 &	67.43 &	\textbf{72.15} \\ \hline
\cite{ghosh2019stance} \small{(msl 128, batch 16, lr 2e-5,
	50e)} & - & - & \textbf{75.10} \\\hline
\end{tabular}
\caption{Overall results for the SemEval 2016 English dataset.}
\label{tab:semeval-results}
\end{table}

For the SVM-based systems, we performed grid search on the training data to
find optimal values for the hyperparameters C and gamma. These can be seen in
Table \ref{tab:svm-parameters}. 

\begin{table}[!ht]
	\centering
	\begin{tabular}{lrrrr} \hline 
	& \multicolumn{2}{c}{\small{\textbf{TF-IDF+SVM}}} & \multicolumn{2}{c}{\small{\textbf{FTEmb+SVM}}} \\ 
		\textbf{Target} & \small{C} & \small{Gamma} & \small{C} & \small{Gamma} \\  \hline 
		Atheism & 700 & 0.001 & 100 & 0.1 \\
		Climate Change & 700 & 0.001 & 10 & 0.75 \\
		Feminist Movement & 700 & 0.001 & 10 & 1.0 \\
		Hillary Clinton & 1000 & 0.0001 & 10 & 0.75 \\
		Legalization of Abortion & 700 & 0.001 & 10 & 1.0 \\ \hline 
	\end{tabular}
\caption{SVM hyperparameters for the SemEval 2016 English dataset.}
\label{tab:svm-parameters}
\end{table}

The fastText system is used off-the-shelf, training domain-specific word
embeddings on each target's training set with one exception: after some
experimentation on the ``Feminist Movement'' train set, we increased the number
of epochs to 60.

The train set from the ``Feminist Movement'' target was also used to obtain the
hyperparameters for mBERT, XLNet, RoBERTa and XLM-RoBERTa. Thus, for
mBERT we used a maximum sequence length of 256, 32 batch size, 5e-5 learning rate
and 5 epochs. 

The results show that, as it has been the case for other text classification
tasks \citep{Devlin19,Yang2019XLNetGA} the Tranformers-based pre-trained language models
outperform any other approach. In fact, we obtain state-of-the art results,
improving over any other approach presented in Section
\ref{sec:related-work}, using the \emph{base} version of RoBERTa
\citep{Liu2019RoBERTaAR}. \cite{ghosh2019stance} obtains best overall results
by applying BERT Large uncased with the hyperparameters specified in the last
row of Table \ref{tab:semeval-results}. This suggests that these large pre-trained language
models require less training data than classic machine learning approaches.
Finally, they also show that for fine-tuning on small datasets it is convenient to increase
the number of epochs.

\subsection{IberEval 2018}\label{sec:results-ibereval2018} 

Tables \ref{tab:result_tw1o_ca} and \ref{tab:result_tw1o_es} report our results
for Catalan and Spanish respectively. The hyperparameters were chosen following
the procedure explained for the SemEval 2016 dataset. The only difference in
the setup refers to the fastText system. While for the SemEval data the word
embeddings were directly trained on the stance training data, for this
particular dataset results were better if we used the fastText pre-trained word
embedding models from Common Crawl \cite{Grave18}. Furthermore, we trained
fastText for 20 epochs.

\begin{table}[!ht]\small
\centering
\begin{tabular}{lccc} \hline
\textbf{System} & \textbf{F1$_{against}$} & \textbf{F1$_{favor}$}  & \textbf{F1$_{avg}$}\\ \hline
TF-IDF+SVM \small{(C=100, Gamma=0.01)} & 22.86 & 94.68 & 58.77 \\
FTEmb+SVM \small{(C=10, Gamma=1)} & 0.00 & 93.88 & 46.94 \\
FTEmb+fastText & 12.90 & 94.60 & 53.78 \\
mBERT \small{(msl 128, batch 32, lr 2e-5, 10e)} & 28.57 & 94.33 & \textbf{61.45} \\
mBERT ca+es \small{(msl 128, batch 32, lr 2e-5, 10e)} & 05.71 & 94.03 & 49.87 \\
XLM-R \small{(msl 128, batch 32, lr 2e-5, 10e)} & 21.62 & 94.86 & 58.24 \\\hline
\textbf{Baseline} \\
\scriptsize{\cite{Cuquerella2018CriCaTM}} & - & - & 30.68 \\  \hline
\end{tabular}
\caption{Results on the TW-1O Catalan testset.}\label{tab:result_tw1o_ca}
\end{table}

\begin{table}[!ht]\small
\centering
\begin{tabular}{lccc} \hline
\textbf{System} & \textbf{1$_{against}$}  & \textbf{F1$_{favor}$} & \textbf{F1$_{avg}$} \\ \hline
TF-IDF+SVM \small{(C=500, Gamma=0.001)} & 68.50 & 64.53 & 66.52 \\
FTEmb+SVM \small{(C=300, Gamma=0.75)} & 63.65 & 58.85 & 61.25 \\
FTEmb+fastText & 69.58 & 65.37 & \textbf{67.48} \\
mBERT \small{(msl 256, batch 32, lr 5e-5, 5e)} & 66.80 & 65.11 & 65.96 \\
mBERT ca+es \small{(msl 256, batch 32, lr 5e-5, 5e)} & 66.67 & 62.16 & 64.42 \\
XLM-R \small{(msl 256, batch 32, lr 2e-5, 10e)} & 65.54 & 59.26 & 62.40\\ \hline
\textbf{Baseline} \\
\cite{Segura-Bedmar18} & - & - & 28.02 \\ \hline
\end{tabular}
\caption{Results on the TW-10 Spanish testset.}\label{tab:result_tw1o_es}
\end{table}

With respect to the results for the Catalan language, it seems to us that its skewed class distribution
is the most important issue, given that every system struggles to correctly
predict the \emph{against} class. The best results are obtained by mBERT, with
a very low 28.57 F1$_{against}$ score. 

For Spanish the results are more balanced. Interestingly, for this language 
the fastText linear classifier combined with fastText embeddings
(FTEmb+fastText) obtains better results than mBERT or XLM-RoBERTa.

We also tried by to augment the training data available to fine-tune the
multilingual pre-trained models by concatenating the training sets of both
languages (e.g., see ``ca+es'' results). However, this strategy was not
beneficial. 

In any case, our results substantially improve over previous state-of-the-art
in both languages. However, it should be noted that they are comparatively
lower than those obtained with the English SemEval 2016 data. In spite of
this, the systems ranking across the English and Spanish datasets is quite
similar, the Catalan results being the exception.

At this point, one question is whether the results
obtained by mBERT and XLM-RoBERTa are lower for Catalan because that language is not as
well represented as English or Spanish in the multilingual language models, or
whether it is just due to the skewed distribution of classes. In the next
sections we will look into these and other research questions using our
CIC corpus.

\subsection{CIC Corpus}\label{sec:cic-results}

The results obtained with both versions of the CIC corpus can be seen in Tables
\ref{tab:result_indep_ca} and \ref{tab:result_indep_es}. Hyperparameters were
chosen on the development set: (i) for TF-IDF+SVM C and Gamma values were 500
and 0.001, respectively; (ii) FTEmb+SVM, C$=$100 and Gamma$=$1; (iii) for fastText
we used the setting described for TW-1O; (iii) mBERT and XLM-RoBERTa were
fine-tuned over 10 epochs using the following settings: maximum sequence length 128, batch 32,
learning rate 2e-5.

\begin{table}[!ht]
\begin{tabular}{lllllll}\hline

 & \multicolumn{3}{l}{\textbf{CIC}} & \multicolumn{3}{l}{\textbf{CIC-Random}} \\ \hline 
\textbf{System} & F1$_{against}$ & F1$_{favor}$ & F1$_{avg}$ & F1$_{against}$ & F1$_{favor}$ & F1$_{avg}$ \\ \hline 
TF-IDF+SVM & 68.89 & 72.91 & 70.90 & 70.20 & 72.49 & 71.35 \\
FTEmb+SVM & 59.43 & 64.46 & 61.95 & 67.91 & 65.77 & 66.84 \\
FTEmb+fastText & 70.73 & 72.21 & 71.47 & 69.62 & 70.63 & 70.13 \\
mBERT & 70.64 & 77.52 & 74.08 & 69.98 & 75.60 & 72.79\\
mBERT ca+es & 53.13 & 77.98 & 65.56 & 60.90 & 77.52 & 69.21 \\ 
XLM-R & 70.69 & 78.67 & \textbf{74.68} & 71.63 & 78.10 & \textbf{74.87}\\ 
\hline
\end{tabular}
\caption{Results on the Catalan testset of the Catalonia Independence Corpus (CIC-CA).}\label{tab:result_indep_ca}
\end{table}

\begin{table}[!ht]
\begin{tabular}{lllllll}\hline
 & \multicolumn{3}{l}{\textbf{CIC}} & \multicolumn{3}{l}{\textbf{CIC-Random}}\\ \hline 
\textbf{System} & F1$_{against}$ & F1$_{favor}$ & F1$_{avg}$ & F1$_{against}$ & F1$_{favor}$ & F1$_{avg}$ \\ \hline 
TF-IDF+SVM & 70.67 & 71.50 & 71.09  & 69.56	& 74.12 & \textbf{71.84} \\
FTEmb+SVM & 64.24 & 62.51 & 63.38 & 64.39 & 66.99 & 65.69 \\
FTEmb+fastText & 73.20 & 71.13 & 72.43 & 69.76 & 71.22 & 70.49 \\
mBERT & 75.17 & 74.27 & \textbf{74.72} & 70.20 & 71.35 & 70.78 \\
mBERT ca+es & 69.54 & 71.83 & 70.69 & 68.02 & 72.86 & 70.44 \\ 
XLM-R & 74.68 & 72.45 & 73.57 & 70.01 & 70.75 & 70.38 \\  \hline
\end{tabular}
\caption{Results on the Spanish testset of the Catalonia Independence Corpus (CIC-ES).}\label{tab:result_indep_es}
\end{table}

\begin{figure}
    \centering
    \includegraphics[scale=0.4]{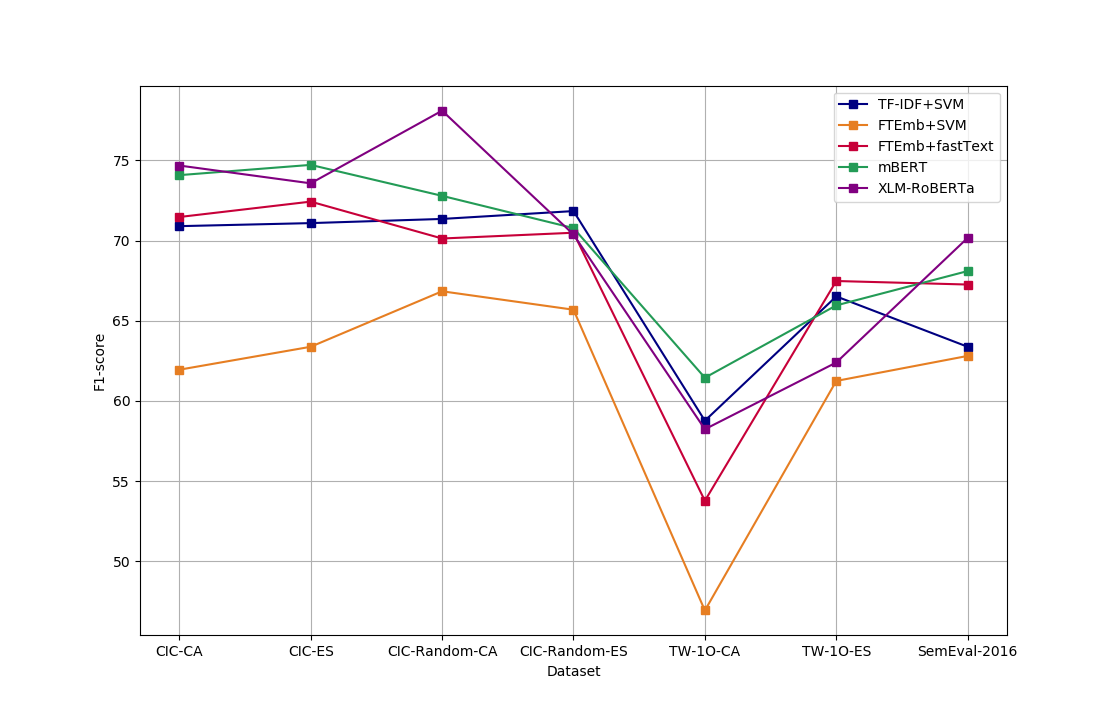}
    \caption{Systems' behaviour across the datasets.}
	\label{fig:systems-behaviour}
\end{figure}

The results from Tables \ref{tab:result_indep_ca} and
\ref{tab:result_indep_es} arise several issues. First, it is clear that there
is a consistency in the behaviour of the systems across both languages. In
fact, it turns out that for the CIC \emph{random} version of the dataset
results are higher in Catalan. Second, systems display similar behaviour across both versions of the CIC dataset. 
This is also true with respect to the
English Semeval 2016 results: the best linear classifier is usually fastText whereas
the Transformer models obtain the best overall results. Therefore, this means that there is not any flaw
in our method for the creation of Twitter-based datasets that may cause the
systems to overfit on the tweets from a specific author's writing style, and so
on. Third, augmenting the training data (``ca+es'') does not help. Finally, both the
Catalan and Spanish results are higher across languages and
systems than those obtained using the TW-1O and SemEval 2016 datasets. These
general patterns are visualized in Figure
\ref{fig:systems-behaviour}. Thus, while for CIC and SemEval the systems display
similar behaviour, with the TW-1O they are much more unstable.

In terms of specific results, it was unexpected that TF-IDF+SVM would
outperform every other approach (in CIC-Ramdom-ES). Furthermore, it is also
surprising that mBERT would obtain better results than XLM-RoBERTa for Spanish;
this was also the case with the TW-1O Spanish
data. In addition, in TW-1O the \emph{favor} class seems easier to learn than \emph{against}.

We believe that the experimental results here presented allows us to conclude that 
our semi-automatic method for the annotation of tweets
presented in this paper is faster and produces annotations 
of comparable quality with the human annotated data at tweet level.

\subsection{Zero-Shot vs Translation}\label{sec:zero-shot-translation}

For our final set of experiments, we apply mBERT and XLM-RoBERTa in a
zero-shot scenario. The idea is the following: assuming that we do not have
annotated data for stance detection in a given language, which would be the
optimal strategy? To answer this question, we consider two alternatives: (i) fine-tune
the models in a given language and predict in the other (zero-shot approach) or, (ii)
fine-tune in the language for which we do have annotated data and predict in
the machine translated version of the other language's test (translation
approach).

\begin{table}[!ht]
\centering
\small
\begin{tabular}{lcccccc}\hline
	& \multicolumn{3}{c}{\textbf{Zero-shot}} & \multicolumn{3}{c}{\textbf{Translate}} \\ 
	& F1$_{against}$ & F1$_{favor}$ & F1$_{avg}$ & F1$_{against}$ & F1$_{favor}$ & F1$_{avg}$ \\ \hline
    \multicolumn{7}{l}{\textbf{CIC Corpus}} \\
	mBERT ca-es & 46.15 & 56.97 & \textbf{51.56}  & 46.51 & 54.75 & 50.63 \\
	mBERT es-ca & 22.07 & 65.47 & 43.77 & 42.92 & 60.73 & \textbf{51.83} \\
	XLM-R ca-es & 48.91 & 56.65 & \textbf{52.78} & 44.80 & 55.69 & 50.25 \\
	XLM-R es-ca & 33.61 & 62.82 & 48.22 & 47.21 & 59.75 & \textbf{53.48} \\ \hline
	\multicolumn{7}{l}{\textbf{CIC-Random Corpus}} \\
	mBERT ca-es & 46.96 & 58.18 & \textbf{52.57} & 45.33 & 56.29 & 50.81 \\
	mBERT es-ca & 33.20 &  60.88 & 47.04 & 50.49 & 55.20 & \textbf{52.85} \\
	XLM-R ca-es & 49.43 & 54.64 & \textbf{52.04} & 43.35 & 57.71 & 50.53 \\
	XLM-R es-ca & 27.87 & 53.10 & 40.49 & 50.75 & 56.58 & \textbf{53.67} \\ \hline
\end{tabular}
\caption{Comparing zero-shot or translating across languages on CIC corpus.}
\label{tab:zero-shots}
\end{table}

Table \ref{tab:zero-shots} reports the result of this experiment. The
\emph{zero-shot} and \emph{translate} multicolumns contain the scores obtained when fine-tuning on a
language and predicting in the other. For example, the first zero-shot row means that
mBERT was fine-tuned on the Catalan CIC training set and evaluated on the CIC-ES
test, obtaining 51.56 F1$_{avg}$ score. Its corresponding \emph{translate}
result means that mBERT was fine-tuned in Catalan and that the prediction was
performed on the translated (into Catalan) CIC-ES test set, obtaining 50.63 F1$_{avg}$
score. For the automatic translation of the target language test sets, we used
the MarianMT system \citep{junczys-dowmunt-etal-2018-marian} via the Hugginface
Transformers API \citep{Wolf2019HuggingFacesTS} which offers MarianMT models
trained on the OpusMT corpus \citep{tiedemann-2012-parallel}.

By looking at Table \ref{tab:zero-shots} we can see that the \emph{translation}
approach works better than applying \emph{zero-shot} whenever the source
language is Spanish, namely, by fine-tuning on Spanish and predicting on the
translated Catalan test set.  We believe that this responds to two claims
already mentioned in the literature. 

First, while these deep learning multilingual models performed very well for
tasks involving high-resourced languages such as English, their performance
drops when applied to low-resource languages \citep{agerri-etal-2020-give}. As
languages share the quota of substrings, which partially depends on corpus
size, larger languages such as Spanish may be better represented than lower
resourced languages such as Catalan. 

Second, it has also been claimed that these multilingual models seem to behave
better for structurally similar languages \citep{karthikeyan2020cross}.

\begin{figure}
    \centering
    \includegraphics[scale=0.5]{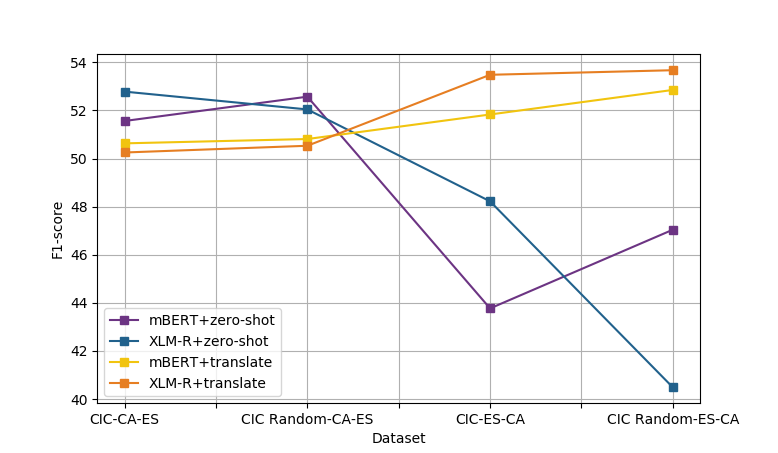}
    \caption{Comparison of zero-shot and translation approaches.}
    \label{fig:zero_shot1}
\end{figure}

Our results reinforce these two separate claims, since the \emph{zero-shot}
approach works quite well when fine-tuning on Catalan and making the
predictions in Spanish (similar languages), and translating is preferable if
the source language is a large and well represented language (such as Spanish).
By plotting the results in Figure \ref{fig:zero_shot1}, it is easier to
confirm this and appreciate two further issues. On the one hand, \emph{translation}
results are more stable because for ``ca-es'' we also fine-tune on the
low-resource language. On the other, the \emph{zero-shot} approach suffers when
predicting onto low-resourced languages. Following this line of argumentation, we also experimented by translating from
Catalan and Spanish into English, but results were substantially worse,
possibly because performance from Catalan to Spanish and viceversa benefits from
their similar grammatical structure.

As a final experiment, we wanted to know how much does machine translation
affect the results reported in Table \ref{tab:zero-shots} and Figure
\ref{fig:zero_shot1}. In order to quantitatively assess this, we fine-tune in a
given language and dataset and evaluate on its translated test set (in a
zero-shot manner). For
example, the first row in Table \ref{tab:loss-translation} means that we
fine-tuned mBERT with the CIC-CA training set and evaluated it on the
translated test set. The setting is the same as in Section
\ref{sec:cic-results}, but making the predictions on the translated test.

\begin{table}[!ht]
\centering
\small
\begin{tabular}{lcccccc} \hline
 & \multicolumn{3}{c}{\textbf{CIC}} & \multicolumn{3}{c}{\textbf{CIC-Random}} \\  
& F1$_{against}$ & F1$_{favor}$ & F1$_{avg}$ & F1$_{against}$ & F1$_{favor}$ & F1$_{avg}$ \\ \hline
	mBERT \textbf{ca-es} & 69.12 & 74.84 & \textbf{71.98} & 68.21 & 70.72 & \textbf{69.47}\\
	mBERT es-ca  & 72.60 & 71.00 & 71.80 & 67.17 & 69.73 & 68.45\\
	XLM-R \textbf{ca-es} & 67.14 & 75.12 & \textbf{71.13} &  67.01 & 73.40 & \textbf{70.21} \\
	XLM-R es-ca & 70.21 & 69.97 & 70.09 & 67.51 & 68.89 & 68.20 \\ \hline
\end{tabular}
\caption{Measuring loss after translation.}
\label{tab:loss-translation}
\end{table}

As it was the case for the \emph{zero-shot} and \emph{translation} comparison,
translating into Spanish produces better results. We attribute this to the
the fact that, while being quite similar, Spanish is a better represented language in the multilingual
pre-trained models. In any case, it is remarkable how good the performance of mBERT and
XLM-RoBERTa is across languages, losing just around 3 points in F1$_{avg}$ score.



\section{Error Analysis}\label{sec:analysis}
After showing in the previous section, by means of quantitative results, that our method generates good quality
annotated data for multilingual experimentation on stance detection in
Twitter, in this section we offer a qualitative analysis to better understand
the features of the Catalonia Independence Corpus (CIC).

In order to do so, we manually revise and annotate a sample from the CIC
training set in order to compare the semi-automatically obtained annotations with
those given by humans. Furthermore, we inspect the output of the best
classifiers, analyzing the tweets that were incorrectly labeled, and calculate
an upperbound score to compare it with our best trained models. 

\subsection{Annotation Errors}\label{sec:errors-dataset}

The manual error analysis of the semi-automatic annotation of the CIC corpus was carried out as follows. We took a random sample of
100 tweets per language from the training sets. The obtained sample was then manually
revised by two human annotators. After independently annotating each of
the samples, the annotators tried to agreed upon a common label for the
contentious examples. Overall, we found out that the error rate in the
Spanish sample was around 20\%, whereas for the Catalan sample was slightly
higher, around 35\%. It should be noted that the annotators found it very difficult to agree on their
correct annotation. This was due to several reasons. 

First, the meaning of the tweets is usually under-specified, namely, there is
not enough context or background information available in order to take an
informed decision. Second, many tweets use figurative language such as sarcasm,
irony or require extra commonsense and/or domain-specific knowledge.  Finally, other tweets
referred to the topic in a indirect manner without clearly establishing a given
stance with respect to the topic.

Table \ref{tab:tweet_err} offers two examples of contentious tweets for which
it is quite difficult to decide whether our semi-automatic method provided a
correct annotation or not. Tweet 1 is classified as NONE in the CIC dataset,
but it is possible to assign both AGAINST (assuming that the writer supports
Arrimadas's action) and FAVOR (if the message was to be a case of irony). In
Tweet 2, although it seems to be of type NONE, a case can be made for it to be
both in FAVOR (assuming that the Spanish judge has an anti-independence bias)
and AGAINST (in this case the tweet would be agreeing with the judge's
decision).

\begin{table}[h]
	\centering
	\footnotesize{
\begin{tabular}{ll} \hline
	\textbf{Tweet 1} & \textit{Arrimadas irá a Waterloo este domingo para
recordar a Puigdemont que} \\
					 & \textit{la república no existe} \\
	\textbf{Our method} & NONE \\
\textbf{Human} & AGAINST/FAVOR \\
\textbf{Language} & Spanish \\
\textbf{Translation} & \textit{Arrimadas will go to Waterloo this Sunday to
remind Puigdemont that} \\
					 & \textit{the Republic does not exist} \\ \hline
\textbf{Tweet 2} & \emph{@unprecisionman @jordisalvia Quan l'advocat preguntava
sobre certes} \\
				 & \textit{contradiccions d'un incident concret q havia
			 explicat el Millo, el jutge} \\
				 & \textit{ha dit q aix\`o no era rellevant per la causa} \\
\textbf{Our method} & AGAINST \\
\textbf{Human} & FAVOR/NONE \\
\textbf{Language} & Catalan \\
\textbf{Translation} & \emph{When the lawyer asked him about certain
contradictions with respect to} \\
					 & \textit{a specific incident which Millo had explained,
				 the judge said that it was} \\
					 & \textit{not relevant.} \\ \hline
\end{tabular}
\caption{Example of a tweet categorized differently with our method and with human annotation.}
\label{tab:tweet_err}
}
\end{table}

This manual annotation exercise showed that labelling stance in tweets is a difficult task for humans, partly
because it depends greatly on the annotator's background knowledge and
intuition. In addition, annotating tweets one by one, as opposed to our user-based
annotation, very often suffers from a lack of context.

\subsection{Prediction Errors}\label{sec:errors-classifiers}

For this analysis we chose the five best classifiers per language from Tables
\ref{tab:result_indep_ca} and \ref{tab:result_indep_es} and identified those
tweets that were incorrectly labeled by at least three of those five classifiers. In total we obtained
134 tweets in Catalan and 125 in Spanish. 

\begin{figure}[!tbp]
  \centering
  \begin{minipage}[b]{0.4\textwidth}
    \includegraphics[width=\textwidth]{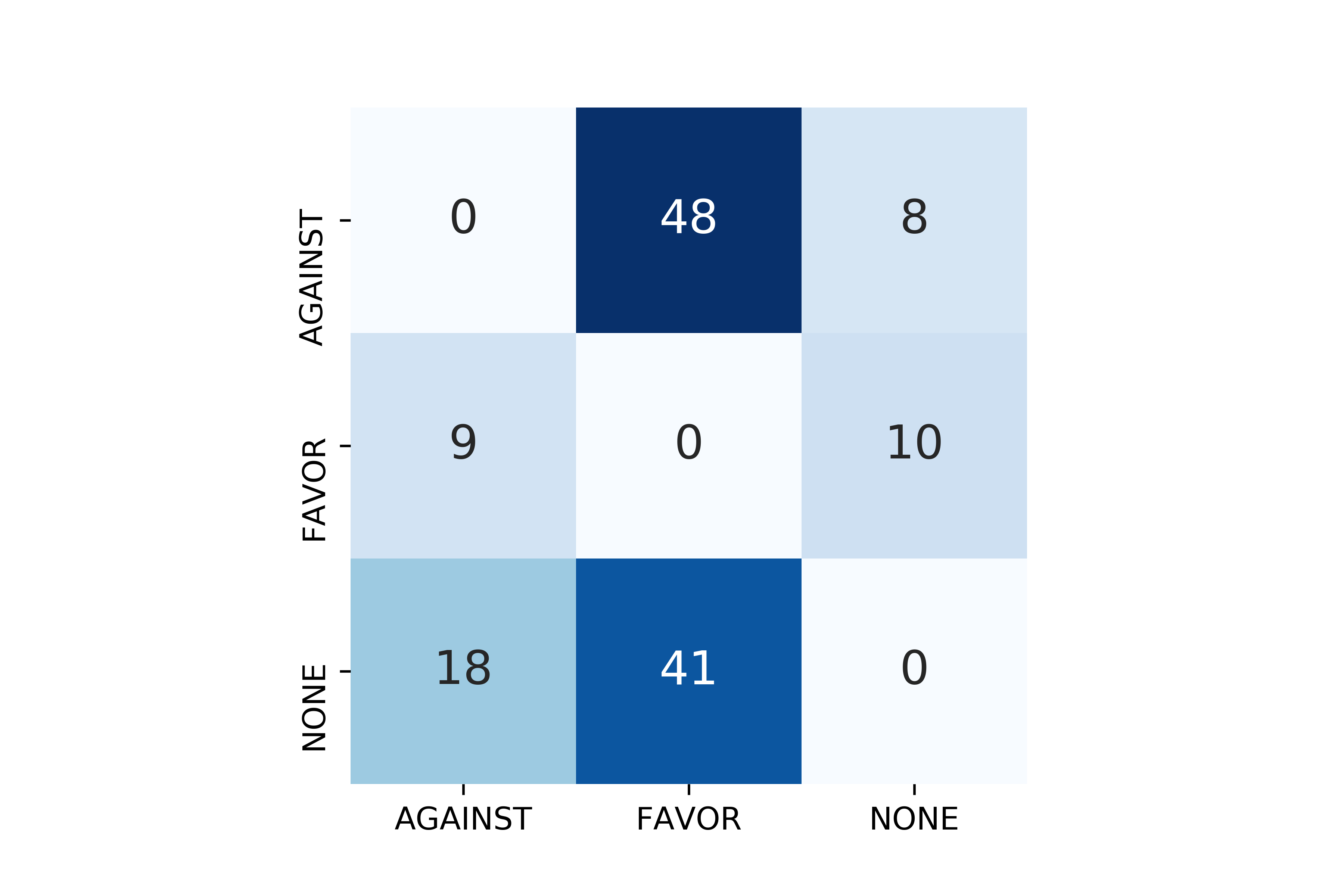}
    \caption{Confusion matrix for CIC-CA made with majority voting}
    \label{fig:errors_ca}
  \end{minipage}
  \hfill
  \begin{minipage}[b]{0.4\textwidth}
    \includegraphics[width=\textwidth]{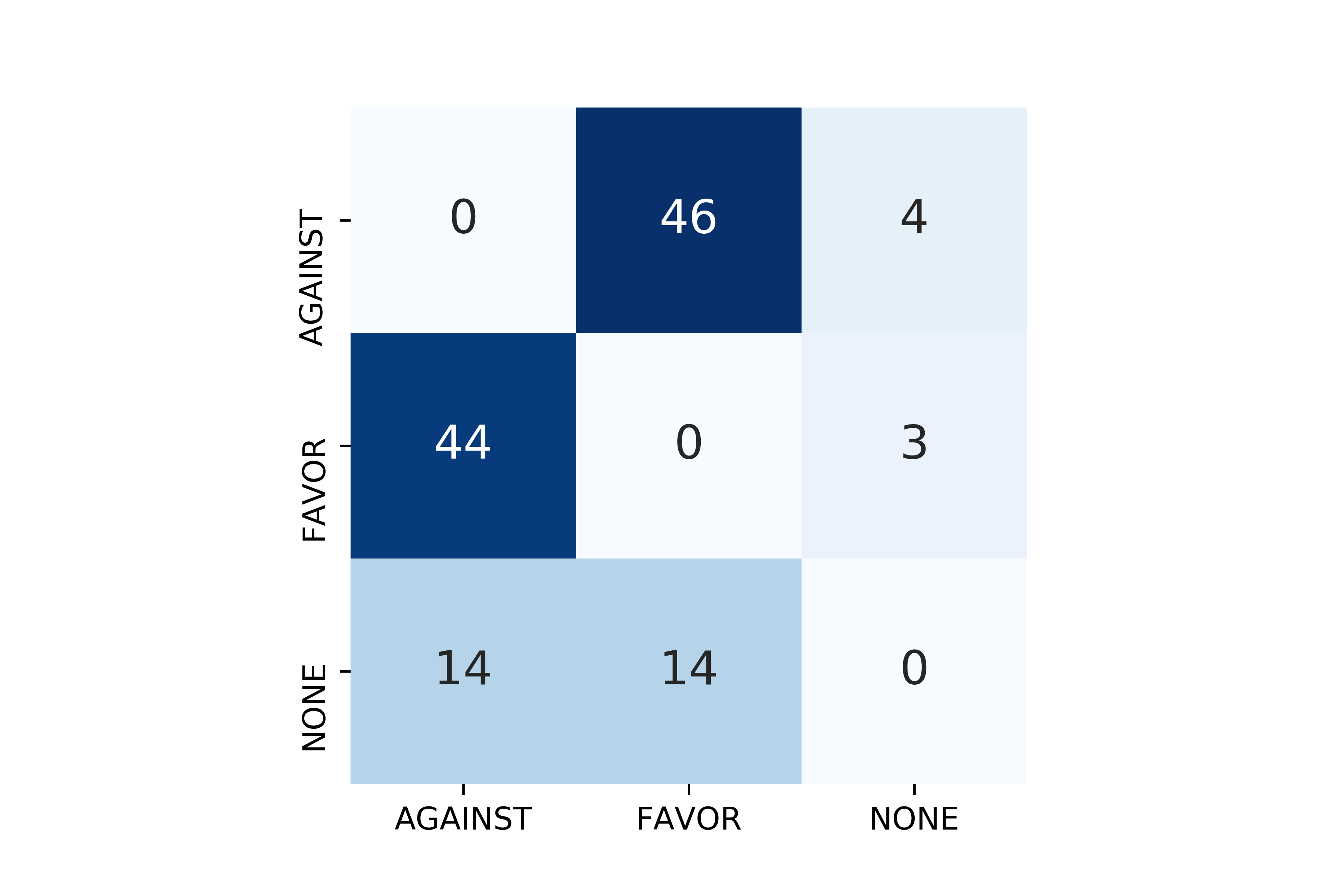}
    \caption{Confusion matrix for CIC-ES made with majority voting.}
    \label{fig:errors_es}
  \end{minipage}
\end{figure}

The type of errors are shown in Figures \ref{fig:errors_ca} and
\ref{fig:errors_es}. It can be seen that the most frequent error for both
languages is AGAINST being predicted as FAVOR. Furthermore, the second most
common source of errors is for Catalan NONE being predicted as FAVOR whereas
for Spanish is FAVOR mistakenly predicted as AGAINST. 

After manual inspection of the misclassified tweets these were the most common
sources of error:

\begin{itemize}
	\item \textbf{Annotation error}: Sometimes when users retweet or quote
		a tweet expressing the opposite political stance without further
		comment. This causes our semi-automatic method to generate wrong
		labels.
    
	\item \textbf{Underspecification}: The tweet is just too short, or the
		target is not explicitly mentioned or referred to.  
	\item \textbf{Missing conversation structure}: Replies to an unknown
		trigger tweet are often difficult to label.
\end{itemize}

Tweet 3 provides an example of a quotation of a message expressing the opposite
stance. In this particular case the message is using content or words related
to an AGAINST stance to express FAVOR. 

\begin{quote}
\textbf{Tweet 3:} \emph{``Joder con los indepes que no se venden como hacía
Pujol''. Voy a informarme de cuál es el grado de cumplimiento de las promesas económicas a los catalanes en general y de los PGE en particular. ¿Por qué no reclamas que te hostien y luego te prometan 4 de tus perras? https://t.co/842QMqkj2W}

\textbf{Label:} FAVOR

\textbf{Automatic classification:} AGAINST (all systems)

\textbf{Language:} Spanish

\textbf{Translation:} \emph{``Bloody independentists who don't sell 
themselves out like Pujol did.'' I'm going to find out what is the degree of
fulfilment of the economic promises made to the Catalans in general and with
respect to the Budget in particular. Why don't you ask to be beaten and then
let them promise you four cents? https://t.co/842QMqkj2W}

\end{quote}

\subsection{Upperbound Score}\label{sec:upperbound}

In order to understand how much room for improvement there is in the CIC dataset, we calculated an upperbound score consisting of assigning a given label if at least
one of the five best systems predicted it correctly. 


Table \ref{tab:upperbound} shows that the gap between the best results and the
upperbound scores is quite large, which means that we still have a large margin
for developing better stance detection systems in this particular dataset. 
\begin{table}[!ht]
\begin{tabular}{lllllll}\hline

 & \multicolumn{3}{l}{CIC} & \multicolumn{3}{l}{CIC-Random} \\ \hline 
\textbf{System} & F1$_{against}$ & F1$_{favor}$ & F1$_{avg}$ & F1$_{against}$ &
F1$_{favor}$ & F1$_{avg}$ \\ \hline 
Upperbound CA & 94.44 & 93.68 & 94.06 & 91.70 & 92.53 & 92.12 \\ 
XLM-R & 70.69 & 78.67 & 74.68 & 71.63 & 78.10 & 74.87\\ \hline
Upperbound ES & 93.63 & 93.71 & 93.67 & 93.23 & 93.67 & 93.45 \\
mBERT & 75.17 & 74.27 & 74.72  & 70.20 & 71.35 & 70.78 \\\hline
\end{tabular}
\caption{Upperbound obtained from the predictions of the best 5 systems.}\label{tab:upperbound}
\end{table}

\section{Conclusion}\label{sec:conclusion}
In this paper we have shown that our methodology to build
annotated datasets for multilingual and cross-lingual stance detection 
in Twitter helps to alleviate a number
of problems present in previous manual-based efforts. Our method to build the
\emph{Catalonia Independence Corpus} (CIC) is faster and
requires less manual effort while obtaining larger and more balanced datasets.
Furthermore, we have empirically demonstrate that the behaviour of the systems
evaluated on the CIC data is consistent with respect to previous benchmarks (e.g., SemEval
2016 and TW-1O). In fact, overall results are higher on the CIC corpus. We have
discarded that those results were due to any overfitting to users'
idiosyncrasies in their writing style by creating the CIC-Random
version of the corpus. Moreover, a qualitative analysis and manual annotation
of a corpus sample showed that the obtained semi-automatic annotations are of
comparable, if not better, quality than the manual ones. We attribute this to
the user-based nature of our method, which helps to overcome the
under-specification or lack of context of the individual tweets.

The availability of the CIC corpus has also allowed us to explore cross-lingual
approaches, comparing zero-shot with translation based strategies. These
experiments have also provided insights
about the behaviour of large pre-trained multilingual language models.

We believe that our method would be helpful to obtain good quality annotated data quicker and
more efficiently for many types of Social Media Analysis and Natural Language
Processing tasks which use Twitter data. In this sense, in future work we would
like to explore the application of our method to various types of tasks
and domains. We see two possible research avenues: (i) on-demand generation of domain-specific
annotated data and, (ii) applying our method to open stance detection, namely, to generate labeled data to
classify the stance of a message not to a specific target, but to a previous
message, news headline or tweet.

\section*{Acknowledgements}

This work has been partially funded by the~Spanish Ministry of Science,
Innovation and Universities (DeepReading RTI2018-096846-B-C21, MCIU/AEI/FEDER,
UE), \textit{Ayudas Fundación BBVA a Equipos de Investigación Científica
2018} (BigKnowledge), and DeepText (KK-2020/00088), funded by the Basque
Government. Rodrigo Agerri is also funded by the RYC-2017-23647 fellowship
and acknowledges the donation of a Titan V GPU by the NVIDIA Corporation.


\end{document}